\documentclass{article}

\PassOptionsToPackage{numbers, compress}{natbib}

\usepackage[final]{tackling_climate_workshop_style}
\usepackage[utf8]{inputenc} 
\usepackage[T1]{fontenc}    
\usepackage{hyperref}       
\usepackage{url}            
\usepackage{booktabs}       
\usepackage{amsfonts}       
\usepackage{nicefrac}       
\usepackage{microtype}      
\usepackage{caption}
\usepackage{graphicx}
\usepackage{amsmath}

\title{Coupling Agent-based Modeling and Life Cycle \\ Assessment to Analyze Trade-offs in \\ Resilient Energy Transitions}

\author{%
  Beichen Zhang \\
  \And
  Mohammed T. Zaki \\
  \And
  Hanna Breunig \\
  \And
  Newsha K. Ajami \\
  \\
  Earth and Environmental Sciences Area\\
  Lawrence Berkeley National Laboratory\\
  \\
  \texttt{\{BZhang5, MTZaki, HannaBreunig, Newsha\}@lbl.gov}
  }
  
\begin{document}

\maketitle

\begin{abstract}
  Transitioning to sustainable and resilient energy systems requires navigating complex and interdependent trade-offs across environmental, social, and resource dimensions. Neglecting these trade-offs can lead to unintended consequences across sectors. However, existing assessments often evaluate emerging energy pathways and their impacts in silos, overlooking critical interactions such as regional resource competition and cumulative impacts. We present an integrated modeling framework that couples agent-based modeling and Life Cycle Assessment (LCA) to simulate how energy transition pathways interact with regional resource competition, ecological constraints, and community-level burdens. We apply the model to a case study in Southern California. The results demonstrate how integrated and multiscale decision making can shape energy pathway deployment and reveal spatially explicit trade-offs under scenario-driven constraints. This modeling framework can further support more adaptive and resilient energy transition planning on spatial and institutional scales.

  
\end{abstract}

\section{Introduction}

The global shift toward resilient energy technologies is not just a technological transformation. From rural deserts to suburban areas, energy transition projects are reshaping land use, drawing on shared water supplies, and influencing the well-being of communities \cite{wu_incorporating_2015, pascale_negotiating_2025}. However, many energy transitions are planned and assessed in silos, leading to unintended consequences of technology deployment and overlooking accumulative impacts on environmental and social dimensions \cite{bolorinos_evaluating_2018, gill_sb_2021}. Currently, Life Cycle Assessment (LCA), Techno-Economic Analysis (TEA), and Integrated Assessment Model (IAM) are widely used to evaluate environmental impacts, economic feasibility, and system-level outcomes of energy transition pathways. While each provides valuable insights, LCA typically operates at fixed scales and relies on static inventory data, and TEA focuses on economic viability but lacks robust impact assessment capabilities \cite{curran_life_2013, yang_chapter_2020, mahmud_integration_2021}. IAMs simulate the deployment of energy technologies with environmental and social impacts, but are generally run at coarse spatial resolutions and large scales \cite{parson_integrated_1997, di_vittorio_e3smgcam_2025}. Therefore, decision-makers lack tools that support resilient energy transitions by capturing local constraints, dynamic trade-offs, and interactive deployment decisions, and account for multiple spatial scales and interdependent environmental and social impacts. 

In this study, we focus on the spatial deployment of emerging transition pathways: green hydrogen production powered by wind or solar, geothermal, waste-to-energy (WtE) with carbon capture, direct lithium extraction (DLE), and direct air capture (DAC), which were selected based on the recent five-year global investment in energy transitions from BloombergNEF datasets \cite{bloombergnef_bloombergnef_2024}. To address the limitations of existing assessment tools and bridge the silos in decision making, we present a novel integrated modeling framework that couples agent-based modeling (ABM) with LCA at regional and site-specific scales. The model simulates the trade-offs and impacts associated with siting and co-location of a portfolio of energy transition pathways, accounting for interactions among pathway and resource constraints, environmental impacts, and community burdens at multiple spatial scales. The paper is organized as follows: Section 2 introduces the modeling framework; Section 3 demonstrates the datasets through the case study in Southern California; Section 4 presents and discusses results; and Section 5 concludes for adaptive energy transition planning and future directions.

\section{Methods and Model Development}

Several approaches have been proposed for combining ABM and LCA, varying in how agents interact with environmental impact assessments. These approaches are broadly categorized as unidirectional and bidirectional coupling \cite{baustert_uncertainty_2017}. Unidirectional coupling includes two variants: ABM enhanced LCA and LCA enhanced ABM. The former has been more commonly used, where ABM is first run to simulate system attributes and decisions, and the resulting outputs are passed to the LCA module \cite{baustert_uncertainty_2017, hicks_seeing_2022}. In contrast, bidirectional coupling, which is adopted in our study, establishes a feedback loop between ABM and LCA, enabling real-time interactions in which decisions and environmental impacts inform each other iteratively \cite{davis_integration_2009, baustert_uncertainty_2017, lan_integrating_2019}.

\begin{figure}[ht]
\begin{center}
\includegraphics[width=13.5cm]{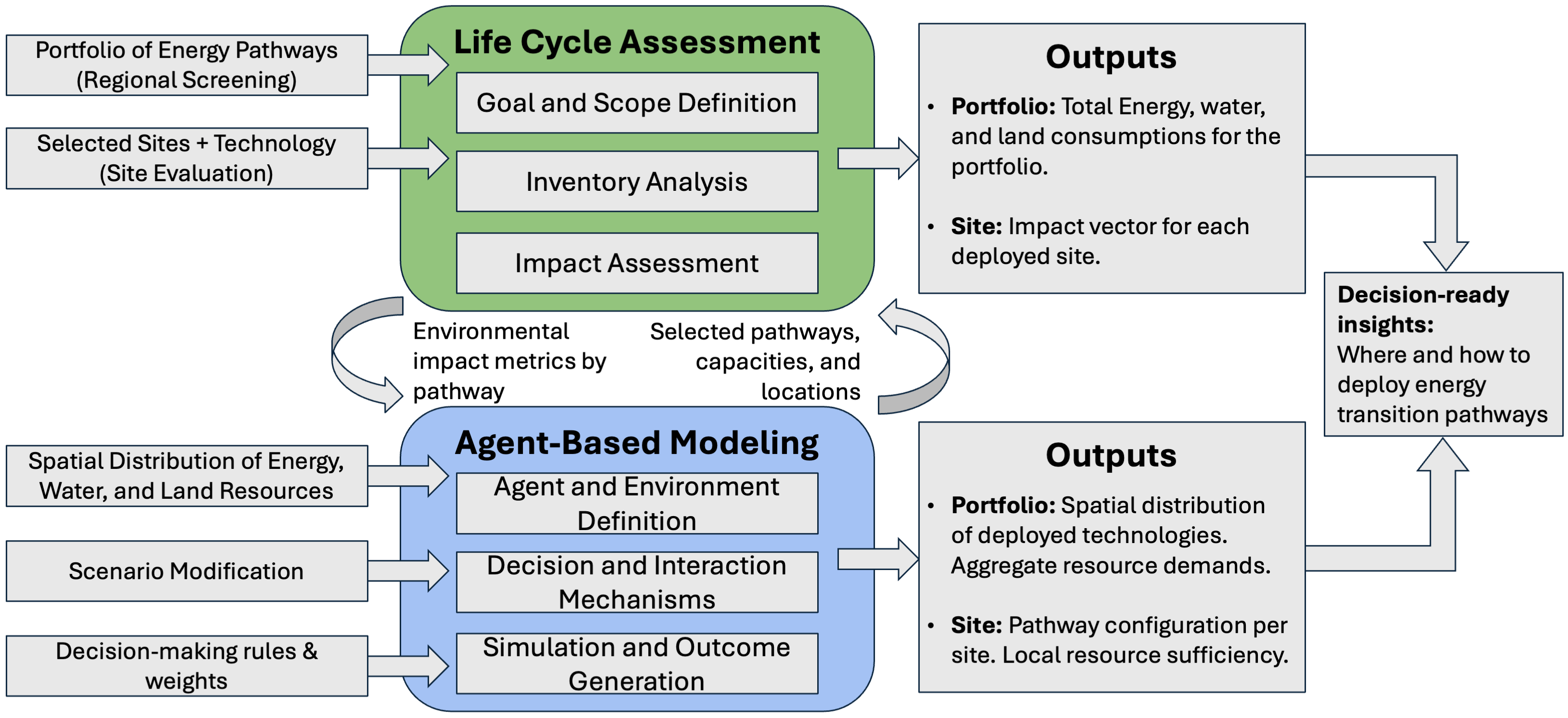}
\end{center}
\caption{Coupling ABM and LCA to model trade-offs in the siting of energy transition pathways and to evaluate environmental impacts at both portfolio and site-specific scales.}
\label{fig:1}
\end{figure}

Figure \ref{fig:1} demonstrates the general framework of the bidirectional coupled ABM-LCA model. At the regional scale, the model begins with a portfolio of proposed energy transition pathways and deployment capacities. The LCA module, following the standard structure, is first applied to calculate total impacts across the portfolio. For a fair cross-comparison of energy pathways, rather than applying LCA across the supply chain, we limit the system boundary to the production step, and the impact assessment focuses on energy, water, and land consumption. This high-level screening provides an initial environmental feasibility check before detailed siting. The updated pathways and their capacities, based on regional resources, are then passed to agents, where the spatial distribution of resources, scenarios, decision-making rules, and weights are also provided to simulate and generate the specific siting configuration. Being dynamically connected, ABM sends the selected pathway, capacity, and location to the LCA module for assessing the environmental impacts at the site level. And the LCA module returns an impact vector to the agents, informing their siting decision. Unsupervised learning-based clustering and multiple-criteria decision analysis (MCDA) are used in the agents to integrate multiscale information and summarize the siting decision based on priorities. In the end, the ABM outputs the spatial distribution of deployed pathways, aggregated demands, and environmental impacts. The ABM-LCA modeling framework offers decision-ready insights on where and how to deploy pathways under competing environmental and social constraints. The comprehensive ABM-LCA workflow (Figure \ref{fig:A1}) and detailed explanation are included in the Appendix.

\section{Model Inputs and Case Study Scenario}

We applied the coupled ABM-LCA model to a case study in Southern California, where ongoing energy transitions interact with environmental stress, climate-related risks, resource competition, and community vulnerability. The model employed a diverse set of multiscale spatial datasets across four major categories: (1) Energy resources, including wind and solar power potential for hydrogen production, geothermal favorability, and waste biomass availability; (2) Water resources, grouped into long-term water transfers, urban water resource, rural water resource, and recycled water; (3) Land resources, covering land use types and area, species biodiversity, terrestrial connectivity, critical habitats, and climate resilience zones; (4) Community vulnerability, derived from features in the revised California Communities Environmental Health Screening Tool (CalEnviroScreen). Additionally, major carbon emission sources and geological carbon storage sites were used to guide DAC siting \cite{westcarb_westcarb_2012}. Existing and newly deployed geothermal power plant locations were also included as siting candidates. A complete list of variables and sources is provided in Appendix Tables \ref{tab:A1} to \ref{tab:A3}. Environmental impact inventories of direct energy, water, and land consumptions for the production stage of each energy pathway were collected from literature reviews. The conversion factors were converted to consistent functional units, summarized in Appendix Tables \ref{tab:A4} to \ref{tab:A7}.

To run simulations, we first defined a portfolio of energy pathways based on diverse real-world cases in Southern California. The portfolio is included in Appendix Table \ref{tab:A8}. In this paper, we applied a baseline scenario that incorporates all resource constraints and community burden indicators. The local constraints for siting in this scenario are based on the resource availability: using only urban water sources and transforming only urban open space or barren land into industrial areas.

\section{Results and Discussion}

\begin{figure}[ht]
\begin{center}
\includegraphics[width=10cm]{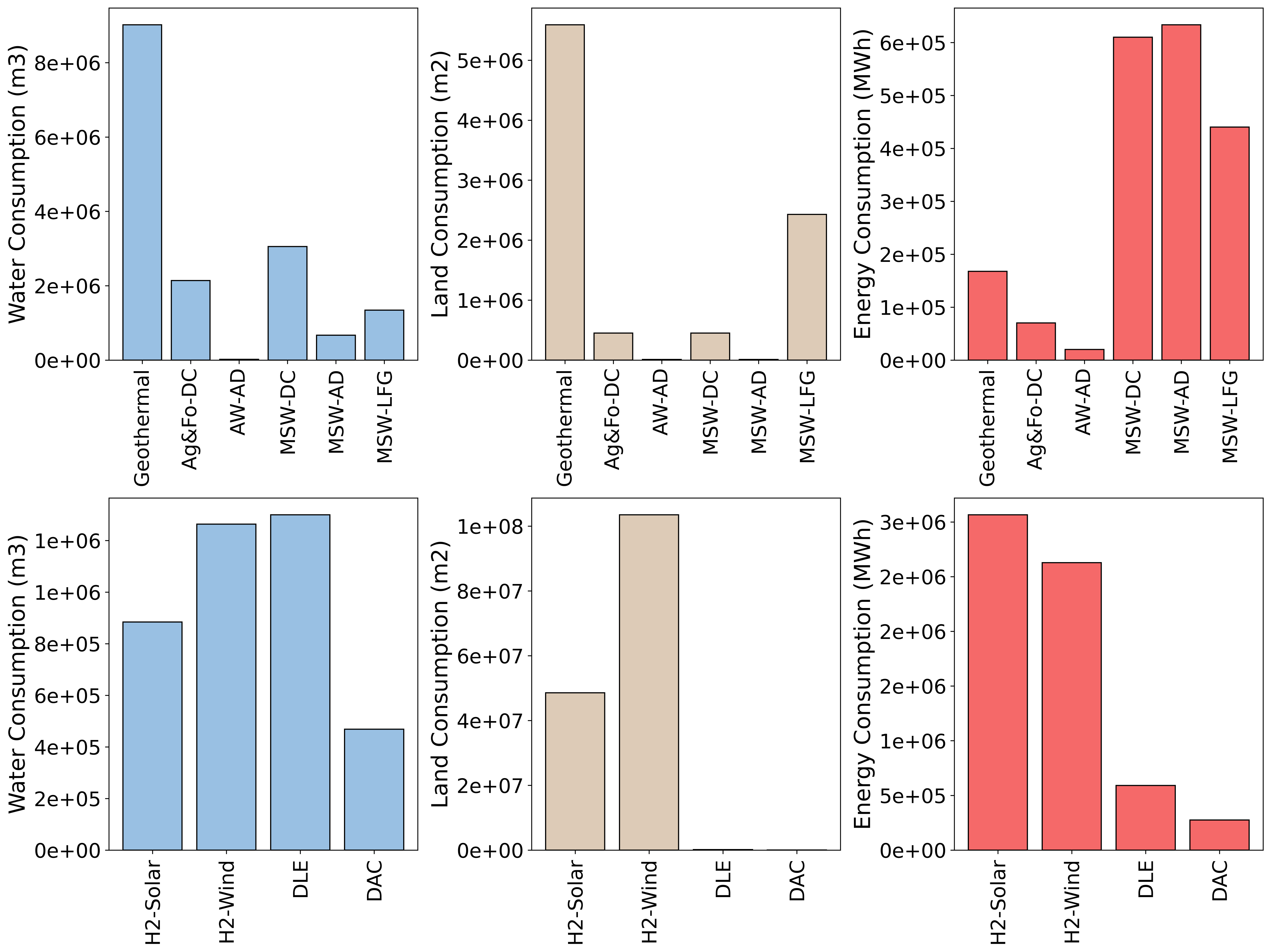}
\end{center}
\caption{Environmental impact assessment for the proposed portfolio in Southern California. Abbreviations: Ag\&Fo, agricultural and forest residues; AW, animal waste; MSW, municipal solid waste; DC, direct combustion; AD, anaerobic digestion; LFG, landfill gas.}
\label{fig:2}
\end{figure}

With the ABM–LCA model fully configured, including the pathway portfolio and baseline scenario for Southern California, the coupled LCA module first evaluated the environmental impacts of each pathway and adjusted their capacities based on the regional resource availability (Figure \ref{fig:2}), before simulating spatial deployment, agent interactions, and interdependent impact assessment. The first row of Figure \ref{fig:2} presents the cumulative impacts from geothermal and WtE technologies, and the second row displays impacts from hydrogen production, DLE, and DAC. The results reveal significant variations in resource demands across pathways. Among electricity generation pathways, geothermal has the highest water and land consumption, and technologies using municipal solid waste (MSW) as the feedstock have the highest energy consumption due to the post-combustion capture requirements. Hydrogen production shows substantial water and land footprints, especially for the wind-powered pathway. Both DLE and DAC require substantial water and are highly energy-intensive. Although they have relatively low land footprints, this analysis includes onsite renewable energy generation for these two pathways, meaning their energy demands impose a substantial burden on the grid.

\begin{figure}[ht]
\begin{center}
\includegraphics[width=13.5 cm]{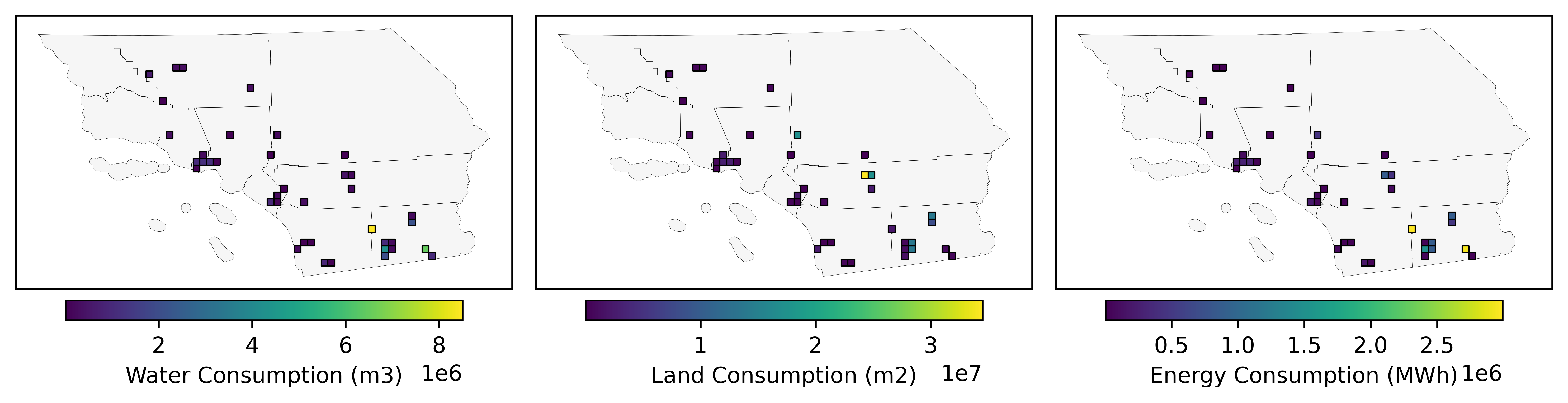}
\end{center}
\caption{Spatial distribution of the cumulative environmental impacts from all deployed pathways.}
\label{fig:3}
\end{figure}

After running the ABM-LCA model under the baseline scenario, all pathways were deployed, except for 8 MW of animal waste with anaerobic digestion for electricity generation and 100 MW of wind-powered hydrogen production, due to local resource constraints and land use limitations. Appendix Figure \ref{fig:A2} demonstrates the spatial patterns of the deployed energy pathways, shaped by the interactions among agents, site-level impact assessments, and local resource availability. Deployed sites are distributed across multiple counties, reflecting the dynamic trade-offs between both local resource feasibility and multi-agent prioritization strategies. Figure \ref{fig:3} presents the cumulative land, water, and energy demands associated with the deployed pathways. The figures reveal how the environmental impacts vary spatially. Local grids show concentrated demand due to either the co-location of multiple pathways or the deployment of high-intensity pathways. For example, land consumption is particularly high in central Riverside County (north of Palm Springs) due to the siting of wind-powered hydrogen production facilities. Meanwhile, water and energy demand peaks in Imperial County due to the deployment of geothermal and DLE pathways. The deployment is primarily located in rural towns and suburban areas, in response to constraints on the water source and land use. The spatial heterogeneity highlights the importance of an integrated and multi-scale management framework that accounts for resource demands, cumulative impacts, and local constraints.

\section{Conclusion}

This study demonstrates a novel integrated modeling framework that couples ABM and LCA to simulate trade-offs and assess impacts on energy, water, and land resources in siting and co-location in energy transition. The interactive and bidirectional loop between the multi-agent system and the LCA module enables the model to capture the dynamic relationships among resource availability, environmental impacts, local constraints, and deployment decisions. The case study in Southern California demonstrates how local resources, social stress, and cumulative impacts influence the siting for a portfolio of emerging energy transition pathways. The results reveal the spatial heterogeneity in deployment patterns and the concentration of cumulative impacts at the local scale. The case study also indicates that even when natural resources and the environment are available at the regional level, local siting is not always successful and can be complicated by co-location and local resource constraints. Our findings support the significance and the necessity of integrated management across resources and pathways at multiple spatial scales for a more resilient energy transition. Future work can focus on improving the adaptive decision-making process and assessing model uncertainty.

\newpage
\bibliographystyle{IEEEtran}

\bibliography{references.bib}
\newpage
\section*{Appendix}

\setcounter{figure}{0}
\renewcommand{\thefigure}{A\arabic{figure}}  

\setcounter{table}{0}
\renewcommand{\thetable}{A\arabic{table}}  

The Shared Resources Pool module centralizes and manages energy feedstock, water supply, and land availability. This module also standardizes the spatial resolution from various polygons and raster datasets to the ten-kilometer grids for data harmonization, enabling cross-resource interactions and dynamic updates throughout the simulation.

The three resource agents (energy, water, and land) interact independently with the Shared Resource Pool and decide the recommended siting based on the ranking of the corresponding resource suitability. Because water and land agents consider multi-scale and multi-dimensional features, we apply unsupervised K-means clustering and a weighted multiple-criteria decision analysis (MCDA) to aggregate the information and rank potential siting locations. In parallel, the Community Burden Agent evaluates the local environmental burdens and risks, incorporating socioeconomic factor and sensitive population indicators to recommend sites with the least community impact.

The siting scores from the four agents, energy, water, land, and community burden, are input to the Integrated Management Agent that performs the weighted MCDA to identify the most suitable deployment location of the energy pathway. The configuration of the energy pathway is then passed to the LCA module for site-specific impact assessment. The impact metrics from the LCA and the location are sent to the Energy Transition Agent, which decides whether to successfully deploy the pathway, scale down the capacity, or relocate to the next best-ranked site from the Integrated Management Agent, based on the impacts, resource availability, and scenario constraints. This decision-making loop of the Integrated Management Agent, Energy Transition Agent, and LCA module enables dynamic and adaptive deployment strategies.

\begin{figure}[ht]
\begin{center}
\includegraphics[width=13.5cm]{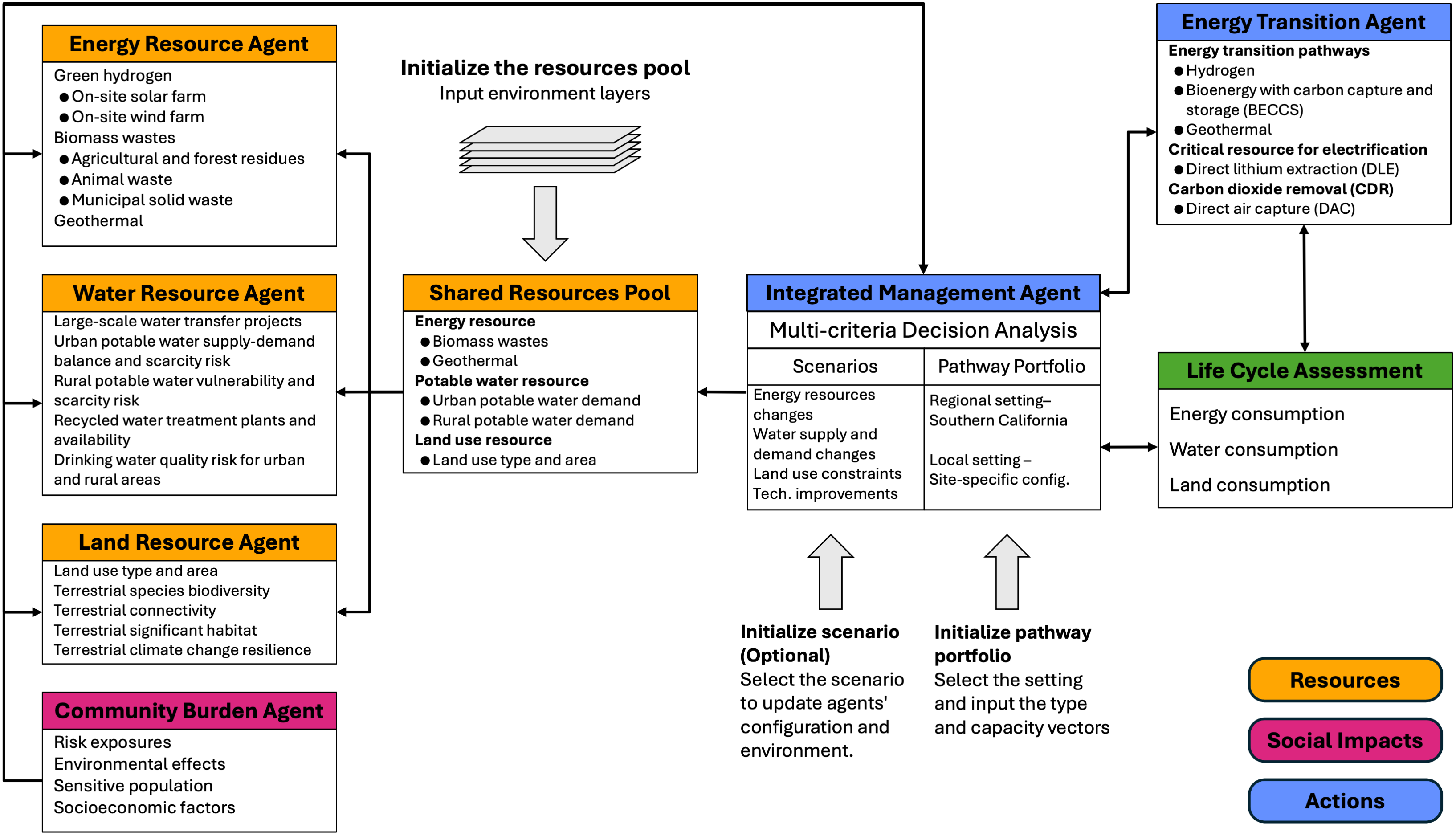}
\end{center}
\caption{The ABM-LCA model workflow illustrating the interactions between agents, LCA module, and environmental layers.}
\label{fig:A1}
\end{figure}

\begin{table}[ht]
\centering
\begin{tabular}{lll}
\toprule
 Category & Variable & Source\\
\midrule
 Hydrogen & Wind potential & Form EIA-860 \cite{eia_electricity_2025}\\

 Hydrogen & Solar potential & Form EIA-860 \cite{eia_electricity_2025}\\
\hline
 WtE & Agricultural and forest residues & Bioenergy KDF \cite{bioenergy_kdf_bioenergy_2024}\\

 WtE & Municipal solid waste & Bioenergy KDF \cite{bioenergy_kdf_bioenergy_2024}\\

 WtE & Animal waste & Bioenergy KDF \cite{bioenergy_kdf_bioenergy_2024}\\
\hline
 Geothermal & Geothermal favorability & USGS \cite{mordensky_when_2023}\\
\bottomrule
\end{tabular}
\vspace{1em}
\caption{Input variables for the Energy Resource Agent.}
\label{tab:A1}
\end{table}

\begin{table}[ht]
\centering
\begin{tabular}{lll}
\toprule
 Category & Variable & Source\\
\midrule
 Long-term water transfer & Number of projects & State water project \cite{california_department_of_water_resources_state_2025}, California \\

 Long-term water transfer & Transfer volume California &  seven-party agreement \cite{us_department_of_the_interior_colorado_2024, us_department_of_the_interior_colorado_2003, palo_verde_irrigation_district_water_1931} \\
\hline
 Urban water & Number of suppliers & California water boards \cite{california_state_water_resources_control_board_electronic_2025}\\

 Urban water & Number of water sources & California water boards \cite{california_state_water_resources_control_board_electronic_2025}\\

 Urban water & Connected to other suppliers & California water boards \cite{california_state_water_resources_control_board_electronic_2025}\\

 Urban water & Water stress index & California water boards \cite{california_state_water_resources_control_board_electronic_2025}\\

 Urban water & Available water volume & California water boards \cite{california_state_water_resources_control_board_electronic_2025}\\
 
 Urban water & Industrial water use ratio & California water boards \cite{california_state_water_resources_control_board_electronic_2025}\\
 
 Urban water & Water quality risk score & California water boards \cite{california_state_water_resources_control_board_drinking_2024}\\
\hline
 Rural water & Number of suppliers & California water boards \cite{california_state_water_resources_control_board_electronic_2025}\\

 Rural water & Number of domestic wells & California water boards \cite{california_department_of_water_resources_drought_2025}\\

 Rural water & Vulnerability of suppliers & California water boards \cite{california_department_of_water_resources_drought_2025}\\

 Rural water & Vulnerability of wells & California water boards \cite{california_department_of_water_resources_drought_2025}\\

 Rural water & Number of water sources & California water boards \cite{california_state_water_resources_control_board_electronic_2025}\\

 Rural water & Connected to other suppliers & California water boards \cite{california_state_water_resources_control_board_electronic_2025}\\
 
 Rural water & Water stress index & California water boards \cite{california_state_water_resources_control_board_electronic_2025}\\
 
 Rural water & Available water volume & California water boards \cite{california_state_water_resources_control_board_electronic_2025}\\
 
 Rural water & Industrial water use ratio & California water boards \cite{california_state_water_resources_control_board_electronic_2025}\\

 Rural water & Water quality risk score & California water boards \cite{california_state_water_resources_control_board_drinking_2024}\\

\hline
 Recycled water & Number of recycle facilities & California water boards \cite{california_state_water_resources_control_board_volumetric_2025}\\
 
 Recycled water & Total flexible capacity & California water boards \cite{california_state_water_resources_control_board_volumetric_2025}\\
 
 Recycled water & Industrial water use ratio & California water boards \cite{california_state_water_resources_control_board_volumetric_2025}\\
\bottomrule
\end{tabular}
\vspace{1em}
\caption{Input variables for the Water Resource Agent.}
\label{tab:A2}
\end{table}

\begin{table}[ht]
\centering
\begin{tabular}{lll}
\toprule
 Category & Variable & Source\\
\midrule
 Land use and land cover & Types and area & USGS NLCD \cite{usgs_annual_2025}\\

 Land use and land cover & Types and area & USGS NLCD \cite{usgs_annual_2025}\\
\hline
 Ecological values & Species biodiversity & California ACE \cite{california_department_of_fish_and_wildlife_areas_2018}\\

 Ecological values & Significant habitats & California ACE \cite{california_department_of_fish_and_wildlife_areas_2018}\\

 Ecological values & Terrestrial connectivity & California ACE \cite{california_department_of_fish_and_wildlife_areas_2018}\\

 Ecological values & Climate resilience & California ACE \cite{california_department_of_fish_and_wildlife_areas_2018}\\
 
\bottomrule
\end{tabular}
\vspace{1em}
\caption{Input variables for the Land Resource Agent.}
\label{tab:A3}
\end{table}

\begin{table}[ht]
\centering
\begin{tabular}{llll}
\toprule
 Pathway & Water Consumption & Unit & Reference\\
\midrule
 Geothermal & 20044.1 & $m^3/MW$ & \cite{dobson_characterizing_2023, clark_water_2011, macknick_review_2011}\\

 Agricultural and forest residues & 30562.6 & $m^3/MW$ & \cite{macknick_review_2011,zhu_water_2019}\\
 
 Animal waste & 4476.6
 & $m^3/MW$ & \cite{macknick_review_2011}\\

 MSW - direct combustion & 303562.6 & $m^3/MW$ & \cite{macknick_review_2011,zhu_water_2019}\\

 MSW - anaerobic digestion & 4476.6 & $m^3/MW$ & \cite{macknick_review_2011}\\

 MSW - landfill gas & 4476.6 & $m^3/MW$ & \cite{macknick_review_2011}\\

 $H_2$ production powered by wind & 25 & $m^3/tonne$ & \cite{ramirez_hydrogen_2023,makhijani_water_2024}\\

 $H_2$ production powered by solar & 15 & $m^3/tonne$ & \cite{makhijani_water_2024} \\

 Direct lithium extraction & 130 & $m^3/tonne$ & \cite{bher_minerals_llc_salton_2020, vera_environmental_2023}\\

 Direct air capture & 3.8 & $m^3/tonne$ & \cite{rosa_water_2021, lebling_6_2022}\\

 Post-combustion Carbon Capture & 2.1 & $m^3/tonne$ & \cite{rosa_water_2021}\\
 
\bottomrule
\end{tabular}
\vspace{1em}
\caption{The conversion factors of the energy transition pathways for assessing the environmental impacts on water supply and demand. \textit{Abbreviation:} MSW, municipal solid waste.}
\label{tab:A4}
\end{table}

\begin{table}[ht]
\centering
\begin{tabular}{llll}
\toprule
 Pathway & Land Consumption & Unit & Reference\\
\midrule
 Geothermal & 12426.5 & $m^2/MW$ & \cite{bayer_review_2013, lovering_land-use_2022}\\

 Agricultural and forest residues & 450000 & $m^2/site$ & \cite{doe_renewable_1997}\\
 
 Animal waste & 10000
 & $m^2/site$  & \cite{fthenakis_land_2009}\\

 MSW - direct combustion & 450000 & $m^2/site$  & \cite{doe_renewable_1997}\\

 MSW - anaerobic digestion & 10000 & $m^2/site$ & \cite{fthenakis_land_2009}\\

 MSW - landfill gas & 2430000 & $m^2/site$  & \cite{vasarhelyi_hidden_2021}\\

 $H_2$ production powered by wind & 345000 & $m^2/MW$  & \cite{lovering_land-use_2022, denholm_land-use_2009}\\

 $H_2$ production powered by solar & 138725 & $m^2/MW$ & \cite{ong_land-use_2013,nrel_life_2012} \\

 Direct lithium extraction & 16 & $m^2/tonne$ & \cite{mousavinezhad_environmental_2024}\\

 Direct air capture & 0.4 & $m^2/tonne$ & \cite{lebling_6_2022}\\

\bottomrule
\end{tabular}
\vspace{1em}
\caption{The conversion factors of the energy transition pathways for assessing the environmental impacts on land use and land cover. \textit{Abbreviation:} MSW, municipal solid waste.}
\label{tab:A5}
\end{table}

\begin{table}[ht]
\centering
\begin{tabular}{llll}
\toprule
 Pathway & Energy Consumption & Unit & Reference\\
\midrule

 $H_2$ production powered by wind & 52 & $MWh/tonne$ & \cite{terlouw_future_2024}\\

 $H_2$ production powered by wind & 52 & $MWh/tonne$ & \cite{terlouw_future_2024}\\

 Direct Lithium Extraction & 59.1 & $MWh/tonne$ & \cite{mousavinezhad_environmental_2024}\\

 Direct Air Capture & 2.2 & $MWh/tonne$ & \cite{lebling_6_2022}\\

 Post-combustion Carbon Capture & 0.4 & $MWh/tonne$ & \cite{jackson_optimization_2019}\\
 
\bottomrule
\end{tabular}
\vspace{1em}
\caption{The conversion factors of the energy transition pathways for assessing the environmental impacts on raw energy demand.}
\label{tab:A6}
\end{table}

\begin{table}[ht]
\centering
\begin{tabular}{llll}
\toprule
 Pathway & Carbon Emission & Unit & Reference\\
\midrule
 Geothermal & 981.1 & $tonne/MW$ & \cite{fridriksson_greenhouse_nodate}\\

 Agricultural and forest residues & 2643.9 & $tonne/MW$ & \cite{mayhead_woody_2020}\\

 Animal waste & 11118.9 & $tonne/MW$ & \cite{eesi_biogas_2017}\\

 MSW - direct combustion & 16057.1 & $tonne/MW$ & \cite{iea_municipal_2003}\\

 MSW - anaerobic digestion & 11118.9 & $tonne/MW$ & \cite{eesi_biogas_2017} \\

 MSW - landfill gas & 3863.7 & $tonne/MW$ & \cite{lee_evaluation_2017}\\
 
\bottomrule
\end{tabular}
\vspace{1em}
\caption{The conversion factors of the pathways of electricity for assessing the carbon emissions. \textit{Abbreviation:} MSW, municipal solid waste.}
\label{tab:A7}
\end{table}

\begin{table}[ht]
\centering
\begin{tabular}{lll}
\toprule
 Pathway & Capacity & Unit \\
\midrule
 Geothermal & 450 & $MW$ \\

 Agricultural and forest residues & 70 & $MW$\\
 
 Animal waste & 10 & $MW$ \\

 MSW - direct combustion & 100 & $MW$ \\

 MSW - anaerobic digestion & 150 & $MW$ \\

 MSW - landfill gas & 300 & $MW$ \\

 $H_2$ production powered by wind & 300 & $MW$\\

 $H_2$ production powered by solar & 350 & $MW$ \\

 Direct lithium extraction & 125000 & $tonne$ \\

 Direct air capture & 10000 & $tonne$ \\

\bottomrule
\end{tabular}
\vspace{1em}
\caption{The proposed portfolio of the energy transition pathways for Southern California. The capacity of each pathway is based on the listed or inferred information from California ARCHES, Bioenergy KDF, Salton Sea geothermal development reports, CTR geothermal lithium project, and California DAC hub.}
\label{tab:A8}
\end{table}

\begin{figure}[ht]
\begin{center}
\includegraphics[width=13.5 cm]{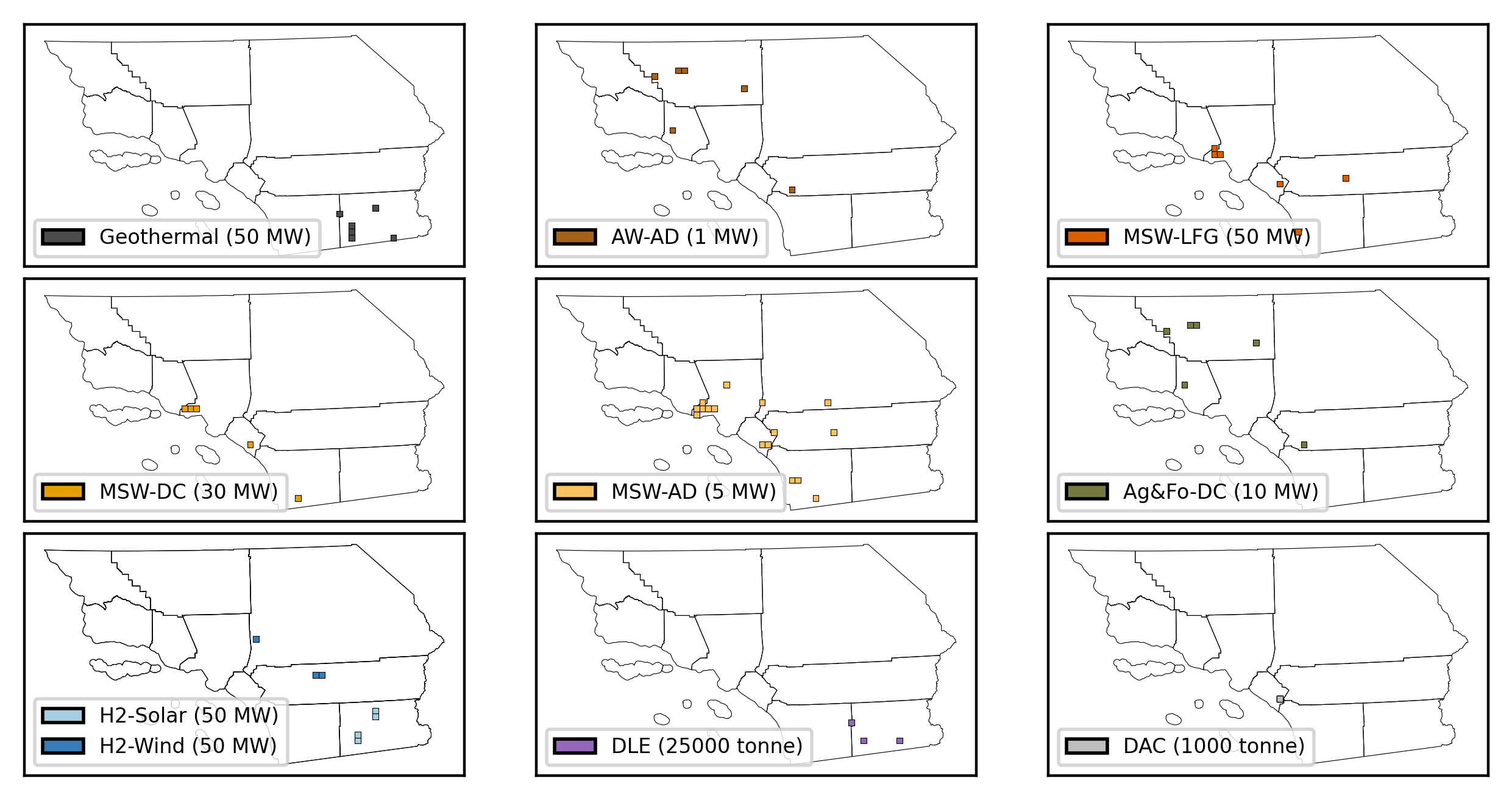}
\end{center}
\caption{Spatial distribution of the deployed energy transition pathways under the baseline scenario.}
\label{fig:A2}
\end{figure}

\end{document}